\documentclass[letterpaper, 10 pt, journal, twoside]{IEEEtran} 

\usepackage[utf8]{inputenc}
\IEEEoverridecommandlockouts                              
\usepackage[dvipsnames]{xcolor}
\usepackage{graphicx}
\usepackage{amsmath,amssymb,amsfonts}
\usepackage{enumerate}
\usepackage{slantsc}
\usepackage{multicol}
\usepackage[ruled,linesnumbered,noend,noline]{algorithm2e}
\usepackage[binary-units=true]{siunitx}
\usepackage[caption=false,font=footnotesize]{subfig}

\newcommand\algname[1]{\textsf{#1}\xspace}
\newcommand{\tstareps}{\algname{T*$\varepsilon$}}
\newcommand{\boldtstareps}{\algname{T*$\boldsymbol{\varepsilon}$}}
\newcommand{\tstar}{\algname{T*}}

\newcommand{\astareps}{\algname{A*$\varepsilon$}}
\newcommand{\astar}{\algname{A*}}

\newcommand{\eps}{$\varepsilon$\xspace}

\newcommand{\nosemic}{\renewcommand{\@endalgocfline}{\relax}}
\newcommand{\dosemic}{\renewcommand{\@endalgocfline}{\algocf@endline}}
\let\oldnl\nl
\newcommand{\nonl}{\renewcommand{\nl}{\let\nl\oldnl}}

\SetKwComment{noNumber}{}{}%

\newtheorem{cor}{Corollary}

\newboolean{HIDENOTES}
\setboolean{HIDENOTES}{false}
\ifthenelse{\boolean{HIDENOTES}}{
    \newcommand{\OS}[1]{{}}
    \newcommand{\JF}[1]{{}}
    \newcommand{\vana}[1]{{}}
    \newcommand{\DP}[1]{{}}
    \newcommand{\StudentInitials}[1]{{}}
    \newcommand{\KS}[1]{{}}
}{
    \newcommand{\OS}[1]{\textcolor{ORANGE}{#1}}
    \newcommand{\JF}[1]{\textcolor{RED}{#1}}
    \newcommand{\DP}[1]{{\color{}#1}}
    \newcommand{\vana}[1]{{\color{RED} #1}}
    
    \newcommand{\StudentInitials}[1]{{\textcolor{BLUE}{#1}}}  
    \newcommand{\KS}[1]{\textcolor{BROWN}{KS: #1}}
}

\newcommand{\ignore}[1]{}
\newcommand{\calT}{\ensuremath{\mathcal{T}}\xspace}
\newcommand{\R}{\mathbb{R}}

\renewcommand{\vec}[1]{\mathbf{#1}}

\usepackage{hyperref}
\hypersetup{
    colorlinks=true,
    linkcolor=black,
    citecolor=black,
    filecolor=magenta,      
    urlcolor=black,
    pdftitle={T*e - Bounded-Suboptimal Efficient Motion Planning for Minimum-Time Curvature-Constrained Systems},
    }

\title{\LARGE \bf
   \boldtstareps---Bounded-Suboptimal Efficient Motion Planning for Minimum-Time Planar Curvature-Constrained Systems
}

\author{Doron Pinsky* \and Petr Váňa$^\dagger$ \and Jan Faigl$^\dagger$ \and Oren Salzman$^\ddagger$
\thanks{Manuscript received: September 9, 2021; Revised December 18, 2021; Accepted January 17, 2022. This paper was recommended for publication by Editor Hanna Kurniawati upon evaluation of the Associate Editor and Reviewers’ comments.}
  \thanks{\textsuperscript{*}Pinsky, D. is with the Technion Autonomous Systems Program
  (TASP), Technion, Haifa 320003, Israel. E-mail: {\tt\small doron.pinsky@campus.technion.ac.il}}
  \thanks{$^\ddagger$Salzman, O. is with the Department of Computer Science, Technion, Haifa 320003, Israel. E-mail:{\tt\small osalzman@cs.technion.ac.il}}
  \thanks{$^\dagger$Váňa, P. and Faigl, J. are with the Faculty of Electrical Engineering, Czech Technical University, 166 27 Prague, Czech Republic. E-mails: {\tt\small \{vanapet1,faiglj\}@fel.cvut.cz}
}
  \thanks{
  This research was supported by grant No.  3-16079 from the Ministry of Science \& Technology, Israel
  and by the Ministry of Education Youth and Sports (MEYS) of the Czech Republic under project No. LTAIZ19013.
}
\thanks{Digital Object Identifier (DOI):  10.1109/LRA.2022.3149307.}
}

\begin{document}

\maketitle

\begin{abstract}
%
We consider the problem of finding collision-free paths for curvature-constrained systems in the presence of obstacles while minimizing execution time.
Specifically, we focus on the setting where a planar system can travel at some range of speeds with unbounded acceleration. This setting can model many systems, such as fixed-wing drones.
Unfortunately, planning for such systems might require evaluating many (local) time-optimal transitions connecting two close-by configurations, which is computationally expensive.
Existing methods either pre-compute all such transitions in a preprocessing stage or use heuristics to speed up the search, thus foregoing any guarantees on solution quality.
Our key insight is that computing all the time-optimal transitions is both~(i)~computationally expensive and~(ii)~unnecessary for many problem instances. 
We show that by finding
bounded-suboptimal solutions (solutions whose cost is bounded by $\boldsymbol{1+\varepsilon}$ times the cost of the optimal solution for any user-provided $\boldsymbol{\varepsilon}$) and not time-optimal solutions, one can dramatically reduce the number of time-optimal transitions used.
We demonstrate using empirical evaluation that our planning framework can reduce the runtime by several orders of magnitude compared to the state-of-the-art while still providing guarantees on the quality of the solution.

\end{abstract}



\section{Introduction}
In this work, we study the problem of finding
minimal-time collision-free paths for curvature-constrained systems with variable speed.
Curvature constraints are prevalent in a~variety of systems~(see,~e.g.,~\cite{needle,autonomous_Ariel,autonomous_Car,garau2009path}).
Unfortunately, determining whether a collision-free curvature-constrained path exists is NP-hard even for a planar system~\cite{NP_hard}.
As a result, this continuous problem can be discretized into a graph data structure using sampling-based~\cite{zeng2015survey} or search-based approaches~\cite{wang2010near}, which is then queried to obtain a discrete path representing a curvature-constrained solution in the continuous space. 
The graph's vertices correspond to robot configurations (i.e.,~$d$-dimensional points that uniquely describe the robot's position and orientation) and edges correspond to local motions taken by the robot.




\begin{figure}[!t]\centering
   \hfill
   \subfloat[][\label{fig:transitions_compare-a}]{\includegraphics[width=0.45\linewidth]{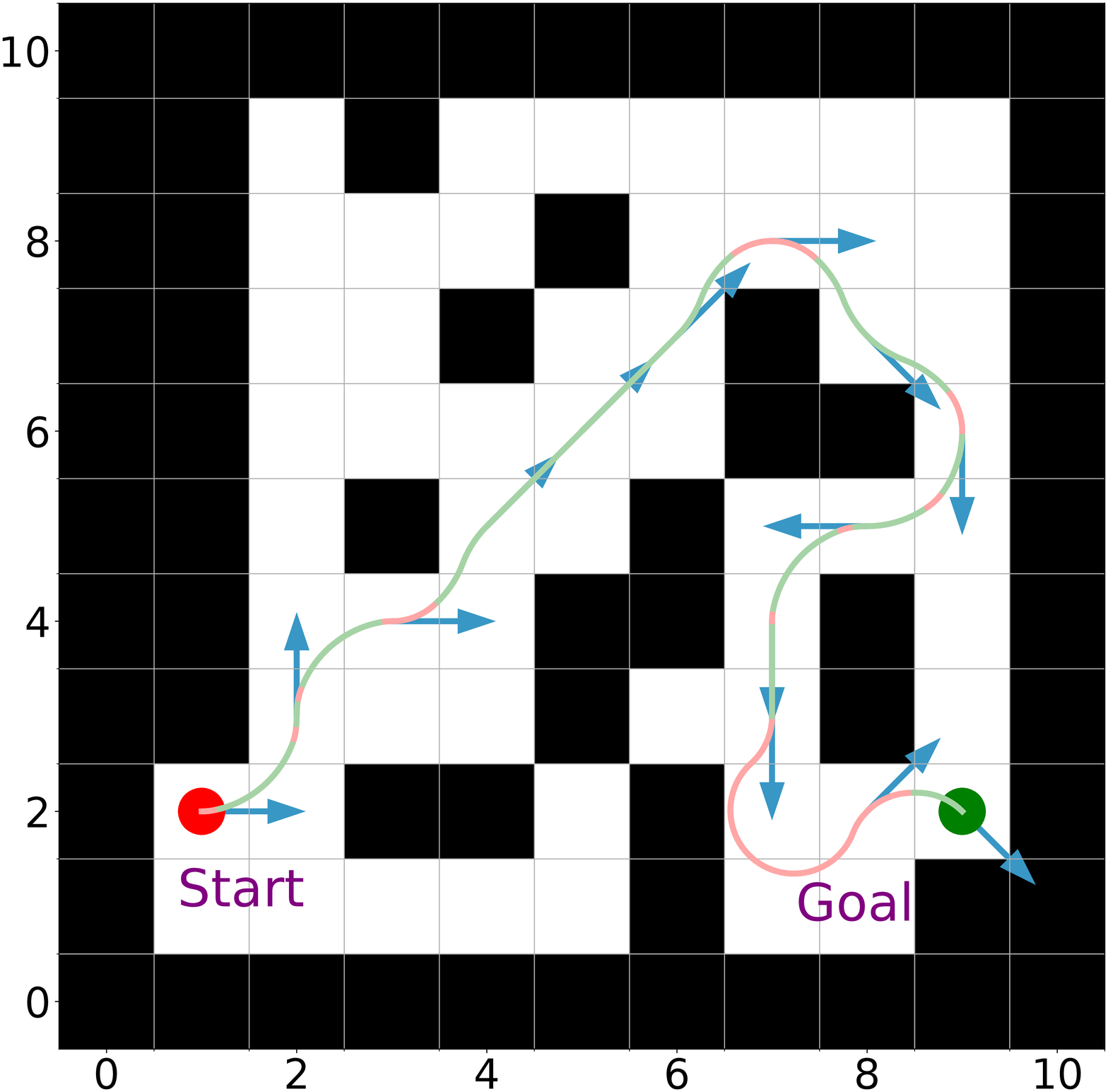}}
   \hfill
   \subfloat[][\label{fig:transitions_compare-b}]{\includegraphics[width=0.45\linewidth]{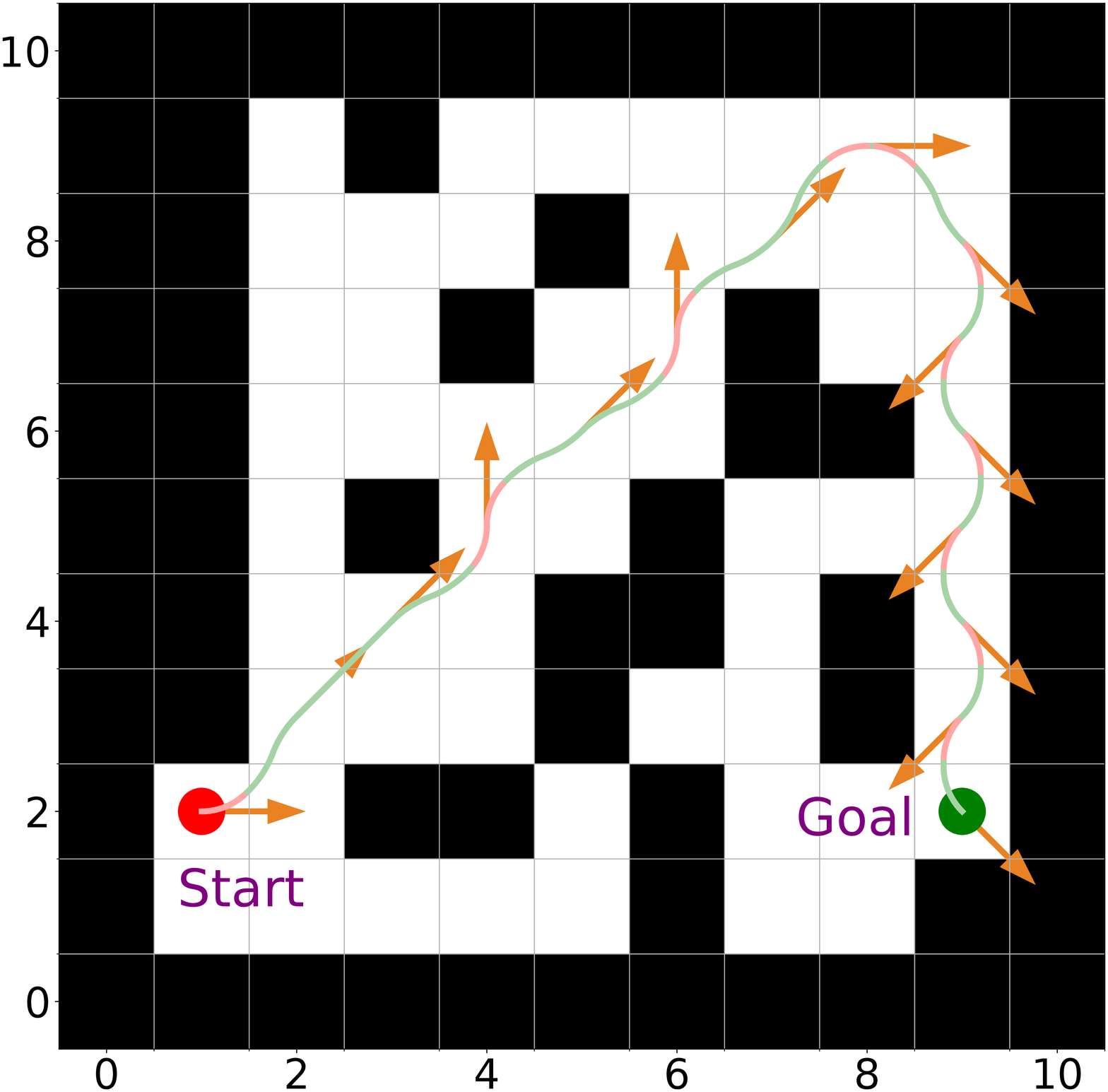}}
   \hfill{}
   \subfloat[][\label{fig:transitions_compare-c}]{\includegraphics[width=0.85\linewidth]{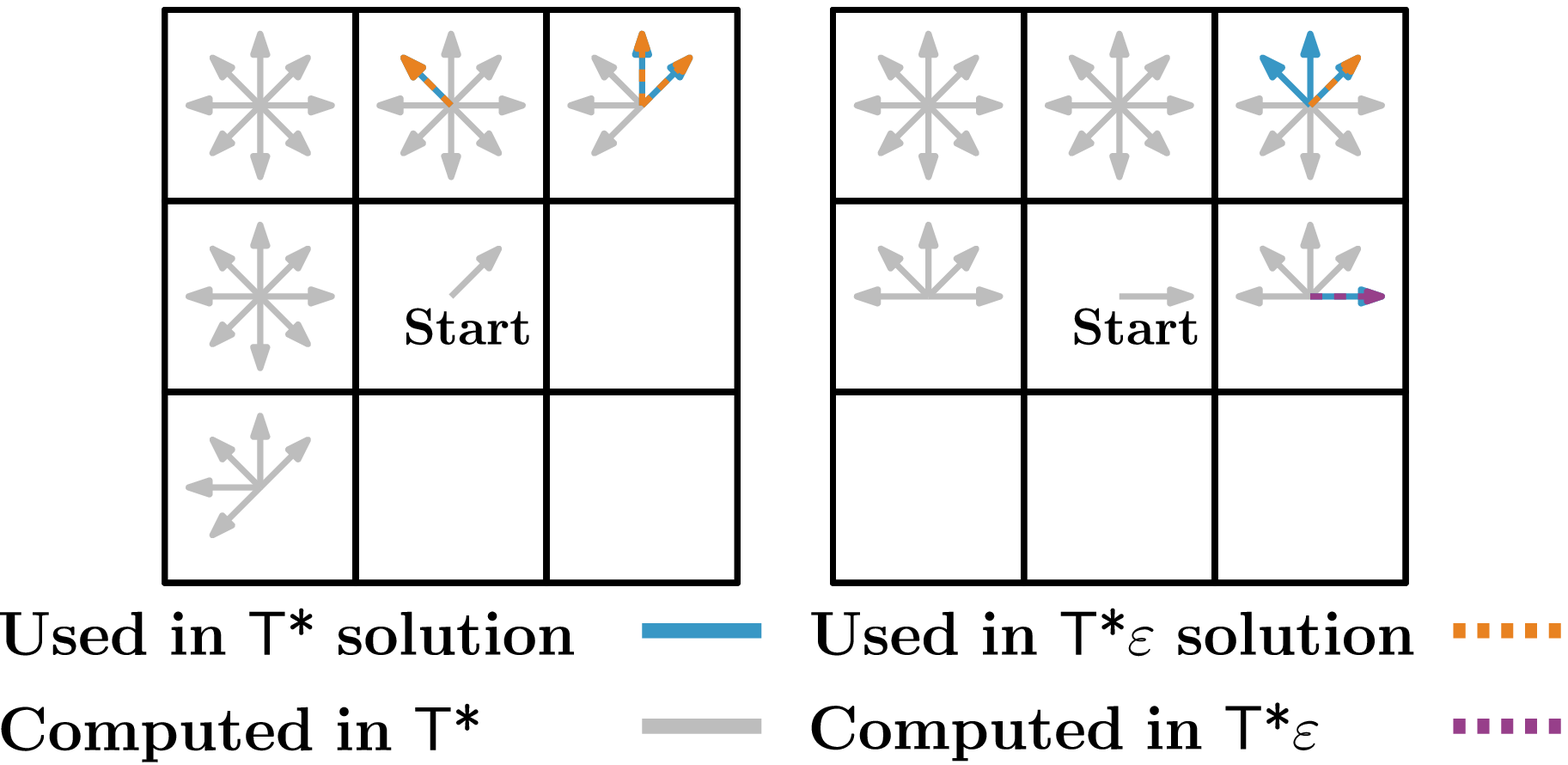}}
   \caption{
      \protect\subref{fig:transitions_compare-a}, \protect\subref{fig:transitions_compare-b}~The time-optimal and bounded-suboptimal path obtained by~\tstar~\cite{T*} and
      by~\tstareps using an approximation factor of~$\varepsilon=2$,~respectively.
      The red and green segments correspond to settings where the system travels at the minimum and maximum speed, respectively.
      The colored arrows show the orientation of each configuration in the solution.
      \tstareps finds a solution whose cost is~\SI{7}{\percent} larger than the cost of the optimal solution found by~\tstar, but the runtime of~\tstareps is faster by a factor of roughly~$13\times$.
      \protect\subref{fig:transitions_compare-c}
     Given a start orientation (diagonal on the left or horizontal on the right) in an eight-connected grid, there are~68 unique transitions in total (after taking into account symmetry and rotation) to the adjacent grid cells.
      {
      The speedup of \tstareps is obtained by computing only a small subset of the time-optimal transitions that are all computed by \tstar.
      Here,~\tstareps computes only five transitions,~four of which are used in the found path.
      Figure best viewed in color.
      }
      \label{fig:transitions_compare}
   }
   \vspace{-5mm}
\end{figure}

Interestingly, the computational bottleneck in these search algorithms is a frequent computation of local time-optimal transitions between neighboring vertices, obtained using numerical optimization.
Our key insight,~depicted in Fig.~\ref{fig:transitions_compare}, is that computing all the time-optimal transitions is both~(i)~computationally expensive and~(ii)~unnecessary for many problem instances. 
We introduce a novel algorithmic framework, which we call~\tstareps, that allows finding
bounded-suboptimal solutions while reducing planning times by orders of magnitude compared to the state-of-the-art.

Our framework consists of three algorithmic components.
The first component assumes that we have an efficient-to-compute~\emph{lower bound} on the cost of the optimal transition between two configurations.
In Sec.~\ref{sec:results}, we demonstrate two such bounds---when the environment does not and does contain dynamics such as wind currents.
The second algorithmic component is the~\astareps-based search algorithm~\cite{astareps} that uses these lower bounds on optimal transitions,~together with a user-provided approximation factor~$\varepsilon$, to choose which edges to consider. 
Roughly speaking,~the search attempts to use only transitions for which the true~(computationally expensive) cost was computed while guaranteeing the bound on the quality of the solution obtained. 
Finally, the third component is a heuristic approach to pre-compute a small set of transitions that are likely to be used in an optimal path.

Using efficient-to-compute lower bounds on the cost of local transitions is a common technique to speed up motion-planning algorithms~(see, e.g., \cite{BK00,H15,DS16,H16,MCSS19}).
These are used to minimize the computation time taken to compute the cost of an edge or transition, also known as an \emph{edge evaluation}.
Typically, edge evaluation corresponds to computing if a robot intersects an obstacle while performing some local motion~(also known as collision detection~\cite{L2006}). 
Thus, the computationally expensive operation of edge evaluation is applied to each edge individually.
Indeed,~it can be shown that under some mild assumptions,~these approaches allow minimizing the number of edges evaluated~\cite{HMPSS18}.

In our case, we plan in a discretized space; thus, the setting is somewhat different.
There is a fixed set of transitions that can be taken from \emph{any} given configuration.
Since these transitions are computationally expensive to compute, it is natural to try and \emph{re-use} transitions that have already been computed.
The challenge is how this should be done while ensuring that the cost of the path found
is within a given multiplicative bound of the cost of an optimal path.

As we demonstrate in simulation (Sec.~\ref{sec:results}), our planning framework allows to compute only a small fraction of transitions, which,~in turn,~allows us to reduce the runtime by several orders of magnitude compared to the state-of-the-art, especially in dynamic conditions.


\section{Related Work}\label{sec:related_work}
We now continue to review related work on planning for minimum-time planar curvature-constrained systems.
When the system is constrained to travel at a single speed, an optimal path connecting two configurations~(i.e., two planar locations, each associated with its angular heading) can be computed analytically~\cite{Dubins}.
In this setting, an optimal path, also known as \emph{Dubins path}, is one of six types:~RSR, RSL, LSR, LSL, RLR, and LRL, where~R and~L refer to right and left turns, respectively, and~S refers to going straight.
As the system travels at a single speed, the shortest and the time-optimal paths are identical.

The kinematic model considered in this work is the setting where the system can travel at some range of speeds but with unbounded acceleration~(i.e., transitioning between low and high-speed can be done instantaneously).
It is not hard to see that, in such a setting, the shortest path is attained by computing the optimal Dubins path, assuming that the system travels at the minimum speed~(as the system can travel using the smallest turning radius possible and can thus better maneuver).
Computing the time-optimal path requires alternating between low-speed ~(to allow for tighter turns) and high-speed motions~(to allow for faster progress).

Wolek \emph{et al.}~\cite{Wolek} showed that it is sufficient to consider only the two extreme speeds and identified a sufficient set of 34 candidate paths (in contrast to the six-candidate paths when the system travels at a single speed).
Each candidate path contains circular arcs and straight-line segments on which the vehicle can travel at either of the two extreme speeds.
However, in contrast to Dubins paths, which can be computed analytically, computing these time-optimal paths for a variable-speed system requires a numerical optimization that is~(i) much more computationally expensive and~(ii) may return locally optimal solutions~(and not globally optimal ones).
Kučerová \emph{et al.}~\cite{kucerova21iros} showed an efficient heuristic approach to find
high-quality paths when considering time as the cost function by using multiple turning radii.
However, there is no guarantee regarding the quality of these paths.

For each model mentioned above, motion-planning algorithms that account for environmental obstacles were introduced.
These include both sampling-based approaches such as the work by Wilson \emph{et al.}~\cite{T*-lite},
search-based methods such as the work by Song \emph{et al.}~\cite{T*}, and hybrid methods that borrow ideas from both motion-planning disciplines~\cite{plaku2013robot}.
Of specific interest to the presented work is \tstar~\cite{T*}, a time-optimal risk-aware motion-planning algorithm that obtains a time-optimal solution but requires a time-consuming preprocessing phase.
As our work builds upon the algorithmic foundation of \tstar, we describe it in Sec.~\ref{sec:alg background}.
Following \tstar, Wilson \emph{et al.}~\cite{GMDM} developed a fast motion-planning algorithm that uses Dubins paths of various speeds to reduce the computational effort and runtime of \tstar. 
For the settings evaluated, the costs of solutions obtained by this algorithm are near-optimal, but there are no guarantees regarding the quality of solutions.

The aforementioned kinematic model can be extended to account for external dynamic changes such as wind or ocean currents.
For such systems, a minimum-time path can be computed similarly to the setting where no wind exists~\cite{minimum_time_const_wind}.
However, only two candidate path types have analytic solutions~(RSR and LSL).
Mittal \emph{et al.}~\cite{Dubins_with_currents} used those two types to compute a path between two vehicle poses under ocean currents.
These kinematic models were used not only in the context of single-goal motion planning problems but also in the settings where the objective is to reach multiple goals~\cite{faigl2017solution} and further extended to account for some notion of rewards~\cite{PFS19}.


\section{Problem Statement}\label{sec:problem}
Our problem formulation follows Song \emph{et al.}~\cite{T*}. 
Specifically, we assume a variable-speed curvature-constrained planar robotic system with unbounded acceleration.
The system's dynamics can be described by
\begin{equation}
   \label{eq:dynamics}
   \left(\begin{matrix}
      \dot{x}(t)\\
      \dot{y}(t)\\
      \dot{\theta}(t)
   \end{matrix} \right)
   =
   \left(\begin{matrix}
      v(t) \cos\theta(t)\\
      v(t) \sin\theta(t)\\
      u(t)
   \end{matrix} \right).
\end{equation}
\noindent 
Here,~$(x,y,\theta) \in \text{SE}(2)$ is the robot's placement and orientation,~$v$ is the robot's speed~(which is considered as a control input) and~$u$ is the second control input dictating the system's turning rate  via maximal lateral acceleration~$K$ that is determined for the specific system\footnote{For example, $K = g \, \tan \varphi_\text{max}$ for fixed-wing vehicles, where $g$ is the gravitational acceleration and $\varphi_\text{max}$ the maximum allowed bang angle.} as
\begin{equation}
   |u| \leq \frac{K}{v}. 
   \label{eq:curvature_limit}
\end{equation}
The speed is limited by some minimum and maximum values denoted by $v_{\min}$ and $v_{\max}$, respectively. W.l.o.g., we assume that~$v_{\max}=1$.
Thus, the system is constrained to travel between two extreme radii:
\begin{equation}
   \rho _{\max}=\frac{v _{\max}^2}{K},
   \quad
   \text{and}
   \quad
   \rho _{\min}=\frac{v _{\min}^2}{K}.
\end{equation}

We assume that the continuous workspace is discretized into a grid according to a predefined resolution. Each cell can be categorized as \emph{free} or \emph{forbidden} corresponding to locations that the system can and cannot occupy, respectively.
In addition, we assume that every robot motion starts and ends at the center of a cell.
Specifically, at each step, the robot transitions to one of its eight neighbors and with one of eight possible orientations. 

A \emph{path}~$\gamma$ is a sequence of configurations where the transition between them obeys the system's dynamics and constraints~(Eq.~\ref{eq:dynamics}).
It is said to be \emph{collision free} if it only occupies free cells.
The cost~$c(\gamma)$ of a path~$\gamma$ is the step-wise cost of the transitions between any consecutive configurations in~$\gamma$. 
It can be computed using the system velocity along~$\gamma$.
Namely,
\begin{equation}
   c(\gamma) = \int_{\gamma} \frac{1}{v(\tau)} d\tau,
   \label{eq:pathCost}
\end{equation}
where~$v(\tau)$ refers to the speed along the path segment~$d\tau$.

Given start and goal configurations~$s_{\text{start}},s_{\text{goal}} \in \text{SE}(2)$, a collision-free path~$\gamma$ is said to be \emph{optimal} if it 
(i)~starts at~$s_{\text{start}}$ and ends at~$s_{\text{goal}}$;
and~(ii)~there is no path~$\gamma'$ connecting~$s_{\text{start}}$ and~$s_{\text{goal}}$ such that~$c(\gamma')<c(\gamma)$.
Similarly, it is said to be \emph{bounded-suboptimal} for some approximation factor~$\varepsilon \geq 0$ if it 
(i)~starts at~$s_{\text{start}}$ and ends at~$s_{\text{goal}}$;
and
(ii)~for any path~$\gamma'$ connecting~$s_{\text{start}}$ and~$s_{\text{goal}}$,~$c(\gamma) \leq (1+\varepsilon) \cdot c(\gamma')$.

While previous works (see, e.g.,~\cite{T*}) were concerned with finding optimal paths, in this work, we are interested in finding
bounded-suboptimal paths for a user-provided approximation factor~$\varepsilon$.
As we will see, the extra flexibility obtained by finding
bounded-suboptimal paths allows us to reduce running times
by orders of magnitude with little compromise on the path quality in practice. 


\section{Algorithmic Background}\label{sec:alg background}

As both \astareps~\cite{astareps} and~\tstar~\cite{T*} serve as the algorithmic foundation of our work, we provide a brief description of these two algorithms.

\subsection{\texorpdfstring{\textsf{A*$\varepsilon$}\xspace} (A-star epsilon)}
\astareps (also known as~\textsc{Focal} search)~\cite{astareps}  is a bounded-suboptimal search algorithm based on  the celebrated \astar algorithm~\cite{astar}.
Like \astar, it searches 
a graph by continuously expanding nodes, starting from the start node until the goal node is reached.
To ensure optimality, \astar orders nodes in a priority list called~\textsc{Open}.
Nodes in the \textsc{Open} list are ordered according to their~$f$-value that is the sum of the cost to reach the node from the source (also known as the cost-to-come or~$g$-value and denoted by $g(n)$ for a node $n$)  added to a conservative estimate of the cost to reach the goal (also known as the cost-to-go or~$h$-value and denoted by $h(n)$ for a node~$n$).
Namely for a node~$n$, its $f$-value is
\begin{equation}
   \label{eq:f-value}
   f(n) = g(n) + h(n).
\end{equation}
Unlike~\astar,~\astareps also uses a so-called~\textsc{Focal} list containing all nodes from the~\textsc{Open} list whose~$f$-values are no larger than~$1 + \varepsilon$ times the smallest~$f$-value in the~\textsc{Open} list. 
It can be shown that expanding any sequence of nodes from the~\textsc{Focal} list ensures that the cost of the final solution will indeed be bounded by a factor of~$1 + \varepsilon$ when compared to the cost of the optimal solution.

\subsection{\tstar (T-star)}
\tstar~\cite{T*} builds on the approach by Wolek \emph{et al.}~\cite{Wolek} that allows finding
an optimal\footnote{
   To be precise, the solutions computed are only an approximation of the time-optimal solutions since the computation is based on numerical optimizations, which may not converge to a global optimum.
   Thus, the optimality of~\tstar and the bounded-suboptimality of~\tstareps is only with respect to (w.r.t.) the quality of the local-optimization.%
   } %
solution for the time-optimal transition between any two configurations in an obstacle-free space.
These transitions are used to compute all possible motions for an agent in a discretized eight-connected grid. 
In this space, there is a finite set of possible transition types.
For example, a transition can be moving from configuration~$(0,0,0)$ to configuration~$(1,1,\pi/2)$, which corresponds to moving diagonally in the NE direction while changing the system's heading.

To this end, there are a total of~$ 8 \times 8 \times 8 = 512$ possible transitions (eight possible start and target orientations and eight neighboring cells).
However, after accounting for symmetry and rotation, there are only 68 unique~transitions. 

After pre-computing all possible transitions, an~\astar-like search is used to find a minimal-cost path.
In their original work, Song \emph{et al.}~\cite{T*} consider both minimal-time paths and a cost-function that balances the time and the risk of colliding with an obstacle.
However, we consider only minimum-time paths in this work and refer to~\tstar as the search algorithm that optimizes this criterion. 

It is important to note that in many settings it is not possible to pre-compute all transitions in an offline phase.
This is because optimization-related parameters might be available only when a query is provided.
For example, in the presence of wind conditions, which affect the system's dynamics and hence the transitions' computation, wind parameters are only available to the planner when a query is provided.
For further details, see Sec.~\ref{sec:results}.

Interestingly, after extensive empirical evaluation, we noticed that: 
(i)~pre-computing all possible transitions last two orders of magnitude more than the \astar-like search;
(ii)~only a small fraction of all possible time-optimal transitions are part of  an optimal path found by~\tstar;
and (iii) Dubins path computed using the minimal speed~$v_{\min}$ (whose cost is a lower bound on the length of the time-optimal path) is often in the same homotopy class as the optimal path and shares a significant portion of its transitions.
These observations are key to our algorithmic framework described in the following section.


\section{Algorithmic framework---\texorpdfstring{\boldtstareps}\label{sec:tstareps}}

\subsection{Preliminaries}
Let~$\calT$ denote a set of possible transitions.
By a slight abuse of notation, we assume that we have access to two functions~$c:\calT \rightarrow \R$ and~$\hat{c}:\calT \rightarrow \R$.
The first, which is expensive-to-compute, corresponds to evaluating the true cost of the transition according to Eq.~\ref{eq:pathCost}.
The second, which is fast-to-compute, corresponds to evaluating a lower bound on the transition cost.
Namely,~$\forall T \in \calT,~\hat{c}(T) \leq c(T)$.
Once the cost of a transition~$T$ is evaluated using~$c(\cdot)$, the value is stored in a cache-like data structure for later use.
Thus, there is no need to compute it again when evaluating the same transition later.

To this end, let~$\kappa:\calT \rightarrow \{0,1\}$ be an indicator function corresponding to the cases where the cost of a transition~$T$~(i)~has not been evaluated using~$c(\cdot)$, i.e.,~$\kappa(T) = 1,$
or~(ii)~has been evaluated using~$c(\cdot)$, i.e.,~$\kappa(T) = 0$.
We start our algorithm with the setting that~$\forall T \in \calT,~\kappa(T) = 1$.\footnote{The reason we chose to define~$\kappa$ in this manner is that we start with~$\kappa(T) = 1$ for all nodes and order nodes lexicographically in the \textsc{Focal} list where lower means better.}


\subsection{Algorithmic Description}

\begin{algorithm}[!t]
  \caption{\text{\boldtstareps}}
  {\label{alg:tstar-eps}}
   \KwIn{$\text{map}, \boldsymbol{s}_\text{start}, \boldsymbol{s}_\text{goal}, \text{approximation factor}\, \varepsilon$}
   \KwOut{bounded-suboptimal path connecting $\boldsymbol{s}_\text{start}\, \text{to}\, \boldsymbol{s}_\text{goal}$}
   \vspace{2mm}
   \noNumber{\color{blue} 
   \texttt{\textbf{S1.}~compute~$c(\cdot)$ for transitions in shortest path }}
   ~$\gamma_{\text{SP}}^{v_{\min}} \leftarrow$  minimal-length collision-free Dubins path on map connecting ${s}_\text{start}$ \text{to}~ ${s}_\text{goal}$ \text{~using}~$v_{\min}$ \nllabel{tstareps-shortest} \\
   \textbf{for each} transition~$T \in \gamma_{\text{SP}}^{v_{\min}}$ \textbf{do} \nllabel{tstareps-for_each_transition}\\ 
   ~~compute~$c(T)$, cache result 
   and set~$\kappa(T) = 0$ \nllabel{tstareps-update_kappa}\\
   \vspace{2mm}
   \noNumber{\color{blue} 
   \texttt{\textbf{S2.}~\astareps-like search}}
   run \astareps from~$n_{\text{start}}$ (node with state  ${s}_\text{start}$) with \textsc{Open} and \textsc{Focal} ordered according to Eq.~\ref{eq:f-value} and~\ref{eq:focal-key}, respectively \nllabel{tstareps-astareps}\\
   \textbf{when} expanding a node~$n$ \textbf{do}\\
   ~~\textbf{if}{ incoming transition~$n.T$ was not computed ($\kappa (n.T) = 1$)} \textbf{then}\\
   ~~~~compute~$c(n.T)$, cache result, set~$\kappa(n.T) = 0$\\
   ~~~~\textbf{continue}\\
   ~~\textbf{if}{ configuration associated with the node~$n$ is~$s_{\text{goal}}$} \textbf{then}\\
   ~~~~\textbf{return} path associated with~$n$\\
   ~~\textbf{for each} succesor node~$n'$ of~$n$ in map \textbf{do} \\
   ~~~~~use Eq.~\ref{eq:g-value} to evaluate the~$g$-value of~$n'$ \nllabel{tstareps-set_g}\\
  \vspace{-1mm}
\end{algorithm}

Our algorithmic framework,
summarized in Alg.~\ref{alg:tstar-eps},
consists of the following two main steps.
\begin{itemize}
    \item[\textbf{S1:}] 
    Heuristically computing the true cost for a small set of transitions.
    \item[\textbf{S2:}] 
    Finding a bounded-suboptimal solution while trying to minimize the number of calls to~$c(\cdot)$.
\end{itemize}

In the first step \textbf{S1} (lines~\ref{tstareps-shortest}--\ref{tstareps-update_kappa}),
we start by finding the shortestcollision-free  Dubins path~$\gamma_{\text{SP}}^{v_{\min}}$~(line~\ref{tstareps-shortest}) from~$s_{\text{start}}$~to~$s_{\text{goal}}$ that uses the minimum speed~$v_{\min}$.
For every transition~$T \in \gamma_{\text{SP}}^{v_{\min}}$ taken along this path, we compute its true cost~$c(T)$, cache it, and set~$\kappa(T)=0$ (lines~\ref{tstareps-for_each_transition}--\ref{tstareps-update_kappa}).

In the second step \textbf{S2} (lines~\ref{tstareps-astareps}--\ref{tstareps-set_g}), we run an~\astareps-like search to find a bounded-suboptimal path from~$s_{\text{start}}$~to~$s_{\text{goal}}$ given some approximation factor~$\varepsilon$.
We base our search algorithm on \astareps because we can use it to heuristically guide the search to expand nodes for which the incoming time-optimal transition has already been computed.

Formally, each node~$n$ considered by the search  
is associated with a parent node~$n.\text{parent}$ except for the start node~$n_\text{start}$ associated with~$s_{\text{start}}$, which has no parent.
In addition, each node stores the time-optimal transition~$n.T$ used to get from~$n.\text{parent}$ to~$n$.
When expanding a node~$n$ (namely, computing successor nodes and inserting them into the \textsc{OPEN} list), 
the true cost~$c(n.T)$ of the time-optimal transition leading to~$n$ is computed.
However, the cost used to reach a successor node~$n'$ via transition~$n'.T$ is 
(i)~$\hat{c}(n'.T)$ if~$\kappa(n'.T) = 1$
and
(ii)~$c(n'.T)$ if~$\kappa(n'.T) = 0$.
Note that for some nodes in the \textsc{Open} list, the true cost of the time-optimal transition associated with them may not have been computed.
However, for all expanded nodes, the true cost of the incoming time-optimal transition is computed at some point.

To this end, if we set~$g(n_\text{start}) = 0$ to be the cost-to-come~($g$-value) of the start node, 
then the cost-to-come of any other node~$n$ in the \textsc{Open} list is
\begin{equation}\label{eq:g-value}
   g(n) = g(n.\text{parent})+ 
   \begin{cases}
      \hat{c}(n.\text{T})  \text{~if~} \kappa(n.\text{T})=1,\\    {c}(n.\text{T})  \text{~if~} \kappa(n.\text{T})=0.
   \end{cases}
\end{equation}

The~$f$-value of the node~$n$ (denoted by~$f(n)$ and used to prioritize nodes in the \textsc{Open} list) is defined  in Eq.~\ref{eq:f-value}.
$h(n)$ can be any admissible estimate of the cost to reach the goal.
In our setting, we use the length of the minimum speed Dubins path divided by the vehicle's maximum speed.

Recall (Sec.~\ref{sec:alg background}) that \astareps uses a \textsc{Focal} list that contains all nodes whose~$f$-value is at most~$1+\varepsilon$ times the~$f$-value of the minimal-cost node in the \textsc{Open} list.
In our setting, nodes in \textsc{Focal} are lexicographically sorted using the following key (from low to high):
\begin{equation}\label{eq:focal-key}
    \text{key}(n) = (\kappa(n.T) , f(n)).
\end{equation}
Thus, any node~$n$ for which~$\kappa(n.T) = 0$ (i.e., the true cost~$c(n.T)$ of the transition leading to~$n$ has been computed) is always prioritized before any node~$n'$ for which~$\kappa(n'.T) = 1$ (i.e., the true cost~$c(n'.T)$ of the transition leading to~$n'$ has not been computed), regardless of their respective~$f$-values.
Among all nodes with identical values of~$\kappa$, nodes with smaller~$f$-values are prioritized.
Once a node~$n$ is chosen for expansion, the true cost of~$n.T$ is computed if it has not been computed beforehand.
Finally, a path is found once a node associated with the goal configuration~$s_{\text{goal}}$ is removed from the \textsc{Open} list.

Following the theoretic properties of \astareps~\cite{astareps} and using the fact that~$\forall T~\hat{c}(T) \leq c(T)$ we have the following Corollary.
\begin{cor}\label{cor:subpot}
    Let~$\varepsilon \geq0$
    and~$\hat{c}(\cdot)$ be some function that bounds from below the true cost~$c(\cdot)$ of any time-optimal transition.
    \tstareps, using~$\varepsilon, \hat{c} \text{, and } c$ is bounded-suboptimal with an approximation factor of~$1+\varepsilon$.  
\end{cor}


\section{Evaluation}\label{sec:results}
In this section, we report on empirical evaluation of our approach in a simulated environment inspired by Song \emph{et al.}~\cite{T*}.
We start~(Sec.~\ref{subsec:eval-S1}) by evaluating our heuristic approach for computing the true cost for a small set of transitions~(step~\textbf{S1}).
Then we continue~(Sec.~\ref{subsec:eval-lb}) to demonstrate how the lower bound~$\hat{c}$ needs to be both informative
(i.e.,~as close as possible to the real cost~$c$) and computationally cheap-to-compute
(as its main purpose is to save  computation time) by evaluating several different lower bounds that balance these two traits. 
We then move to compare our motion-planning approach with~\tstar 
in static environmental conditions~(Sec.~\ref{subsec:eval-static}), i.e., when the system dynamics are known before the query is provided and are independent of the specific location at which the robot resides,
and
in dynamic environmental conditions such as when wind currents exist~(Sec.~\ref{subsec:eval-dynamic}).
Finally, we briefly discuss and qualitatively compare~\tstareps with alternative approaches to find
paths that attempt to minimize the execution time (Sec.~\ref{subsec:additional_approaches}).

All experiments were run on a \SI{3.1}{\giga\hertz} Intel Core i9 processor with \SI{32}{\giga\byte} of memory.
Benchmarks, system parameters, and C++ implementation of all the algorithms used in our work are publicly available\footnote{ \url{https://github.com/CRL-Technion/tstar-epsilon}.}.
{Throughout Sec.~\ref{subsec:eval-S1}--\ref{subsec:eval-static}, we test the performance of our algorithm across~$100$ randomly generated scenarios of $14 \times 14$ cells and using~$v_{\min}=0.5$, where for a given scenario, each cell has a probability of~\SI{25}{\percent} of being blocked, and the start and goal configurations are chosen uniformly at random while ensuring that a solution exists.
When reporting algorithmic attributes, e.g., solution cost or runtime, the values represent the average across the~$100$ scenarios. 
In some cases, we also report confidence intervals that include two standard deviations from the mean.  
In Sec.~\ref{subsec:eval-dynamic}, 
we provide additional information on the experiment setup.}



\subsection{Evaluating the Efficacy of step \textbf{S1}}\label{subsec:eval-S1}

\begin{figure}[t!]\centering
   \includegraphics[width=0.6\linewidth]{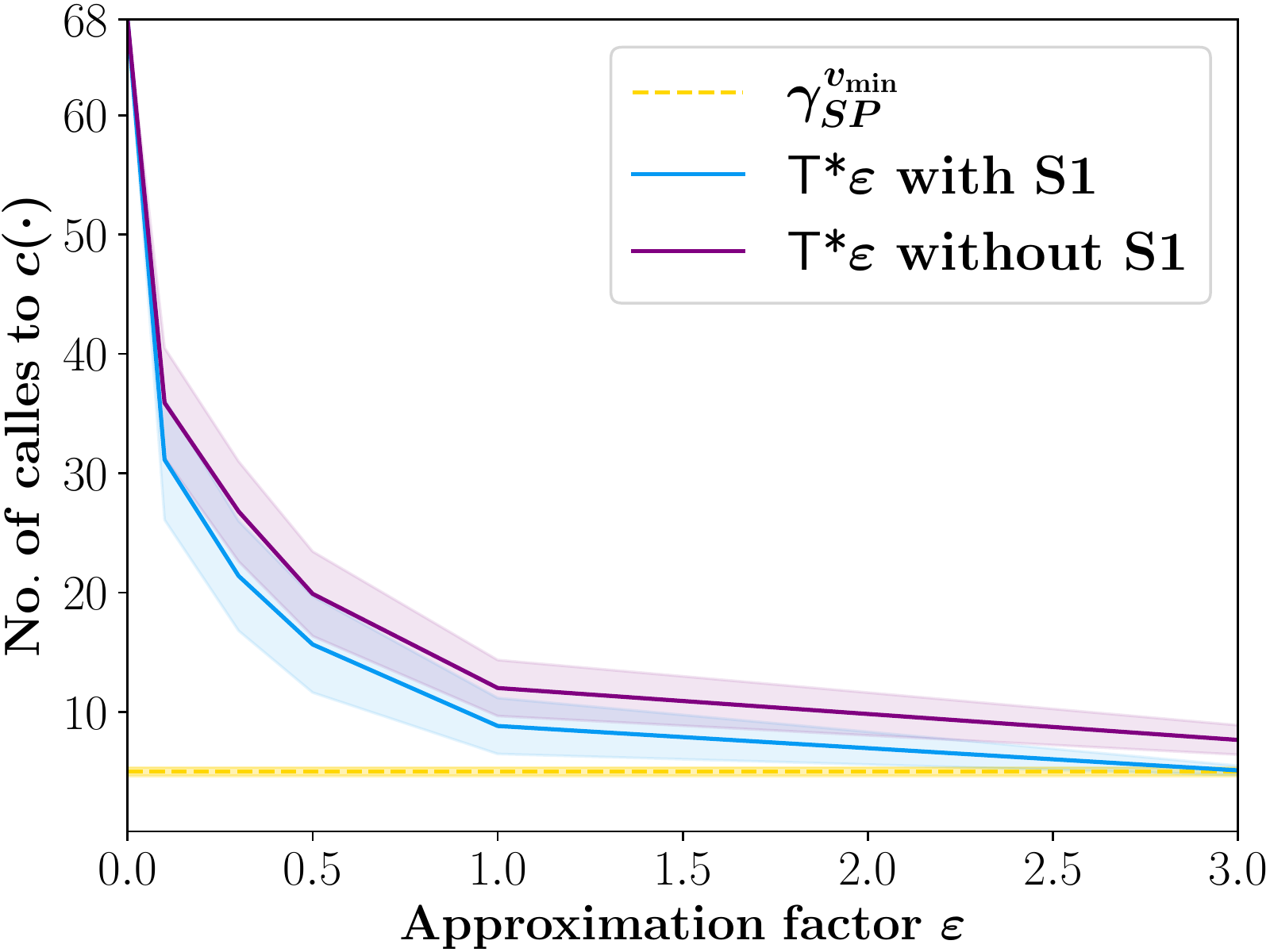}
   \caption{
   {Average number of calls to $c(\cdot)$ 
   by \tstareps  as a function of~$\varepsilon$ with (blue) or without (purple) step \textbf{S1},
    respectively. 
    Recall that there are at most~$68$ possible transitions.}
   \label{fig:total_transitions}
   }
   \vspace{-2mm}
\end{figure}
To evaluate the efficacy of the step \textbf{S1}, wherein we provide a heuristic approach for computing the true cost for a small set of transitions, we report the number of calls to~$c(\cdot)$ 
by \tstareps as a function of the approximation factor~$\varepsilon$ with and without this initial step (Fig.~\ref{fig:total_transitions}).


{Observe that for values of~\eps larger than $0.1$, using  \textbf{S1} allows reducing the average number of calls to~$c(\cdot)$ by  roughly~$10\%$.}
Moreover, in this case, for large approximation factors, the average calls to~$c(\cdot)$ converges to the number of transitions in~$\gamma_{SP}^{v_{\min}}$.
%
When the step \textbf{S1} is not invoked, \tstareps is not ``bootstrapped'' with a good set of pre-computed time-optimal and thus evaluates unnecessary transitions.

\subsection{Computing Lower Bounds---Balancing Computation Time with Informative Lower Bounds\label{subsec:eval-lb}}

Recall that in Sec.~\ref{sec:tstareps}, we assumed that we have access to~$\hat{c}$---an efficient-to-compute tight lower bound on~$c$, the true cost of the time-optimal transition.
However, there is an inherent tradeoff between how informative a lower bound is (which immediately corresponds to how accurately it can guide the search)
and its computation time
(which can negate the effectiveness of the lower bound).

We evaluated \tstareps using three different lower bounds that demonstrate this behavior:
(i)~Euclidean distance divided by~$v_{\max}$;
(ii)~length of minimum-speed Dubins path divided by~$v_{\max}$;
and
(iii)~length of minimum-speed Dubins path divided by~$v_{\max}$ while accounting for environment obstacles.
Note that all lower bounds are divided by~$v_{\max}$, thus always underestimating the true cost of a transition. In addition, they are ordered from low to high according to their computation time and how informative they are.

\begin{figure}[!t]
\captionsetup[subfigure]{labelformat=empty}
\centering
    \vspace{-5mm}
   \hfill
   \subfloat[][\label{fig:lowerBounds-a}]{\includegraphics[width=0.9\linewidth]{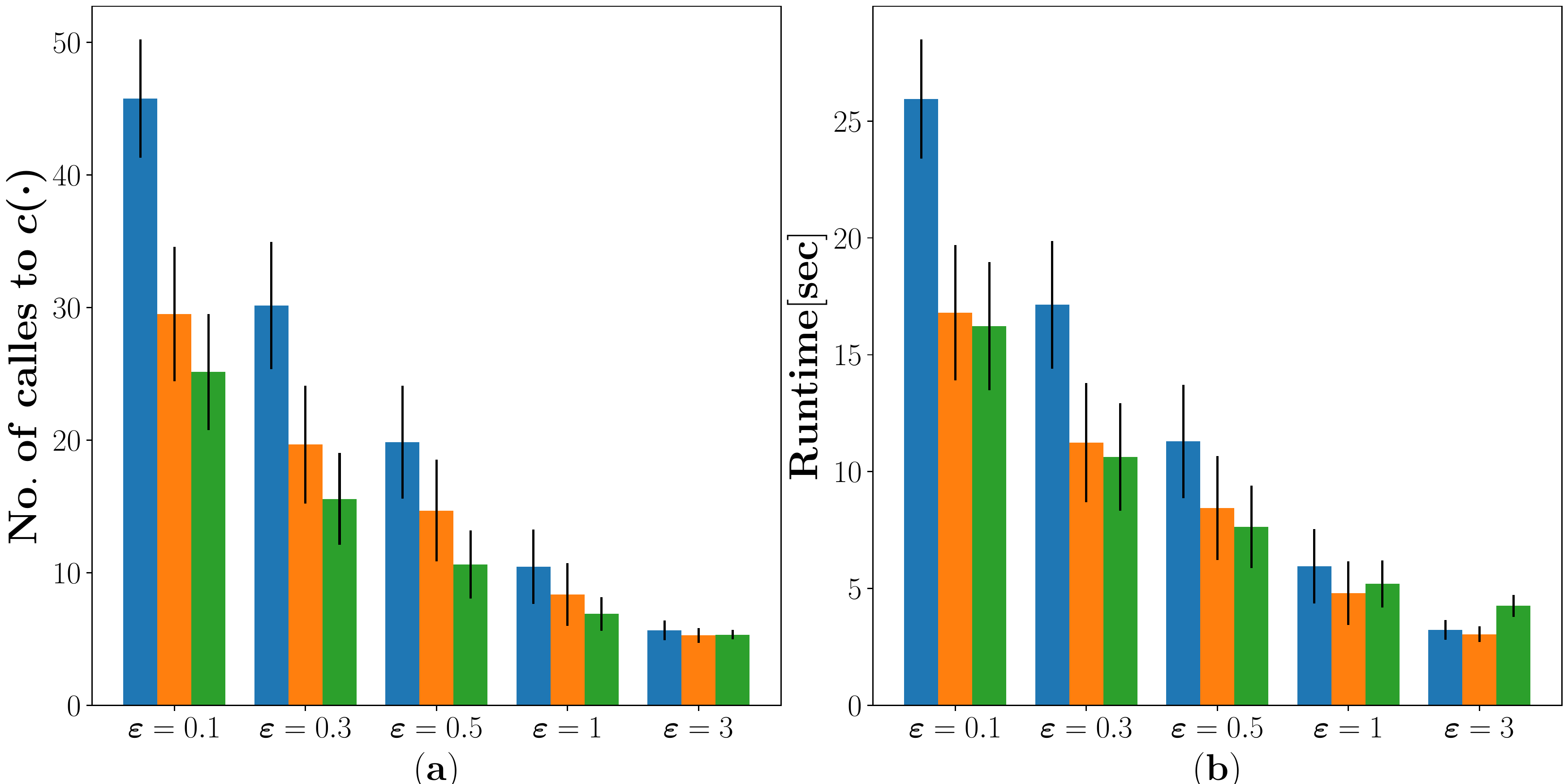}}
   \hfill{}
   \vspace{-8mm}

   \subfloat[\label{fig:lowerBounds-c}]{\includegraphics[width=1\linewidth]{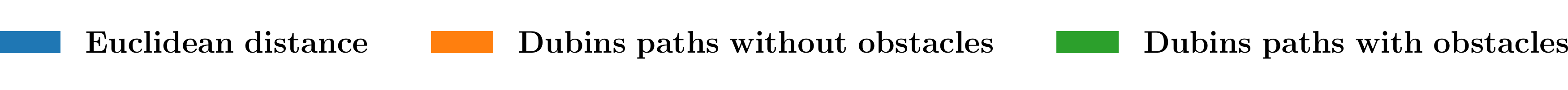}}

   \caption{
   {Comparing the effect of different approaches for lower bounds ($\hat{c}(\cdot)$ in Eq.~\ref{eq:g-value}) {for several values of the approximation factor~\eps}.
   (\hyperref[fig:lowerBounds]{a})~Average number of calls to~$c(\cdot)$
   and
   (\hyperref[fig:lowerBounds]{b})~Average running time of~\tstareps, respectively.}
   }
   \label{fig:lowerBounds}
   \vspace{-5mm}
\end{figure}



\begin{figure*}[t]\centering
  \subfloat[\label{fig:no_wind_results-a}]{\includegraphics[height=3.5cm]{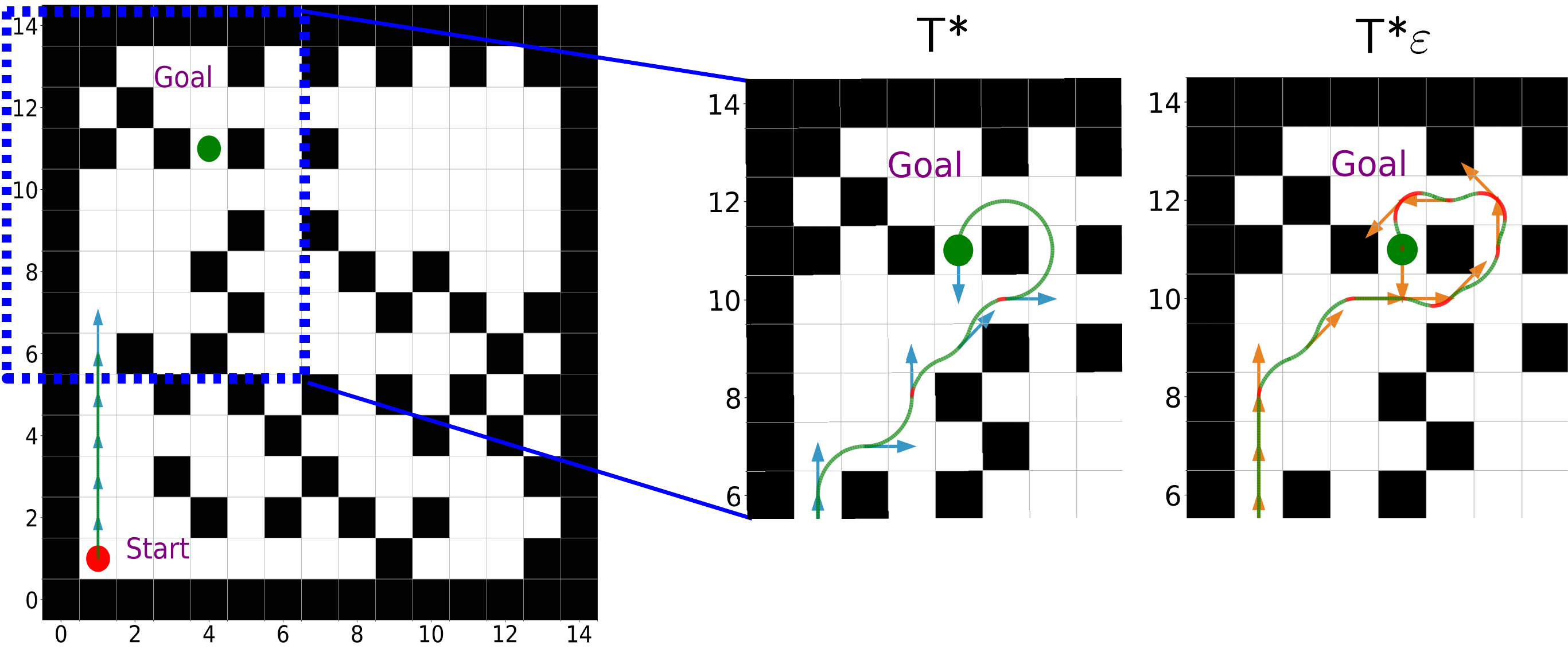}}
   
  \hfill{}
   
   \subfloat[\label{fig:no_wind_results-b}]{\includegraphics[height=3.8cm]{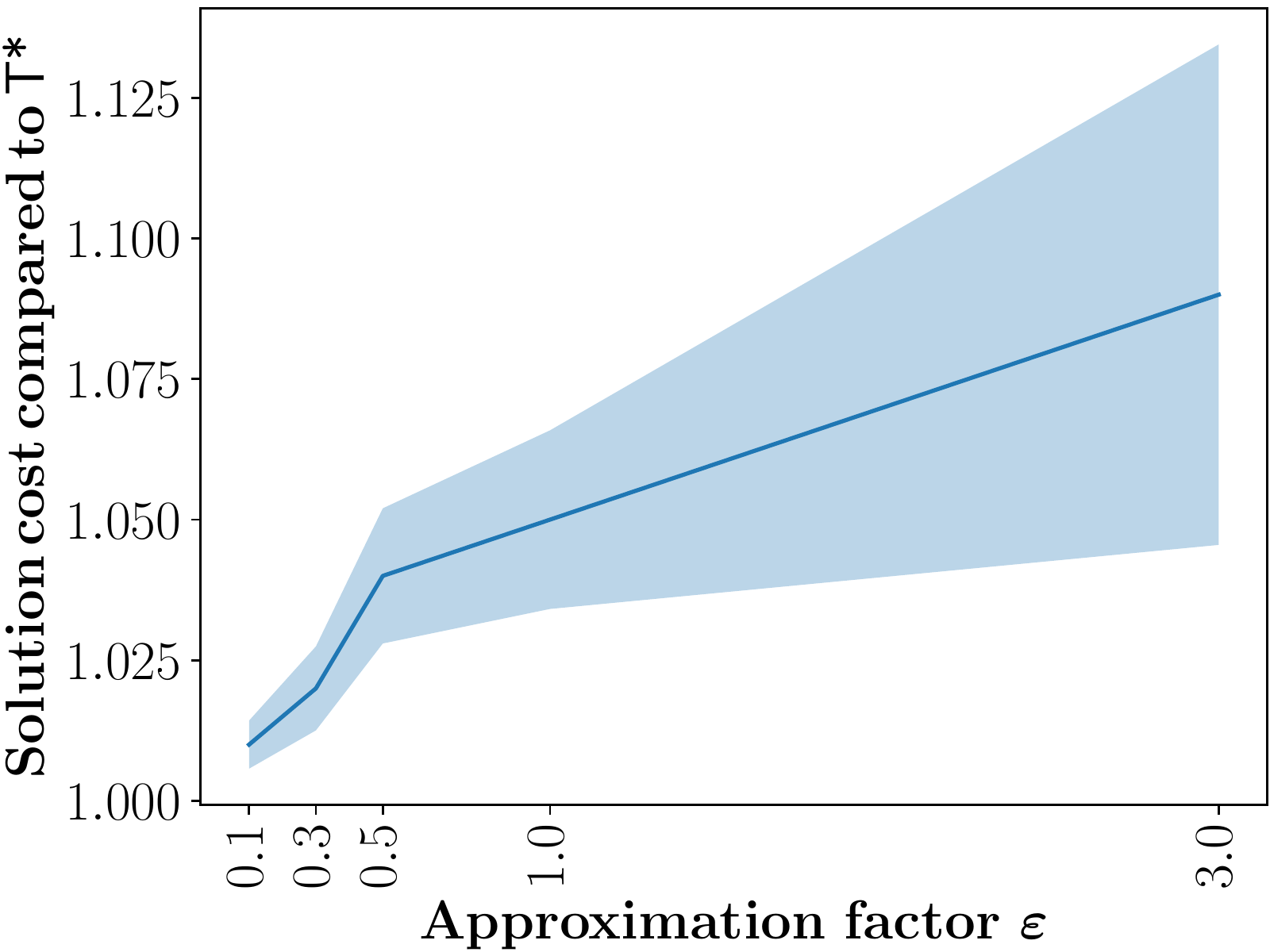}}\,
   \subfloat[\label{fig:no_wind_results-c}]{\includegraphics[height=3.8cm]{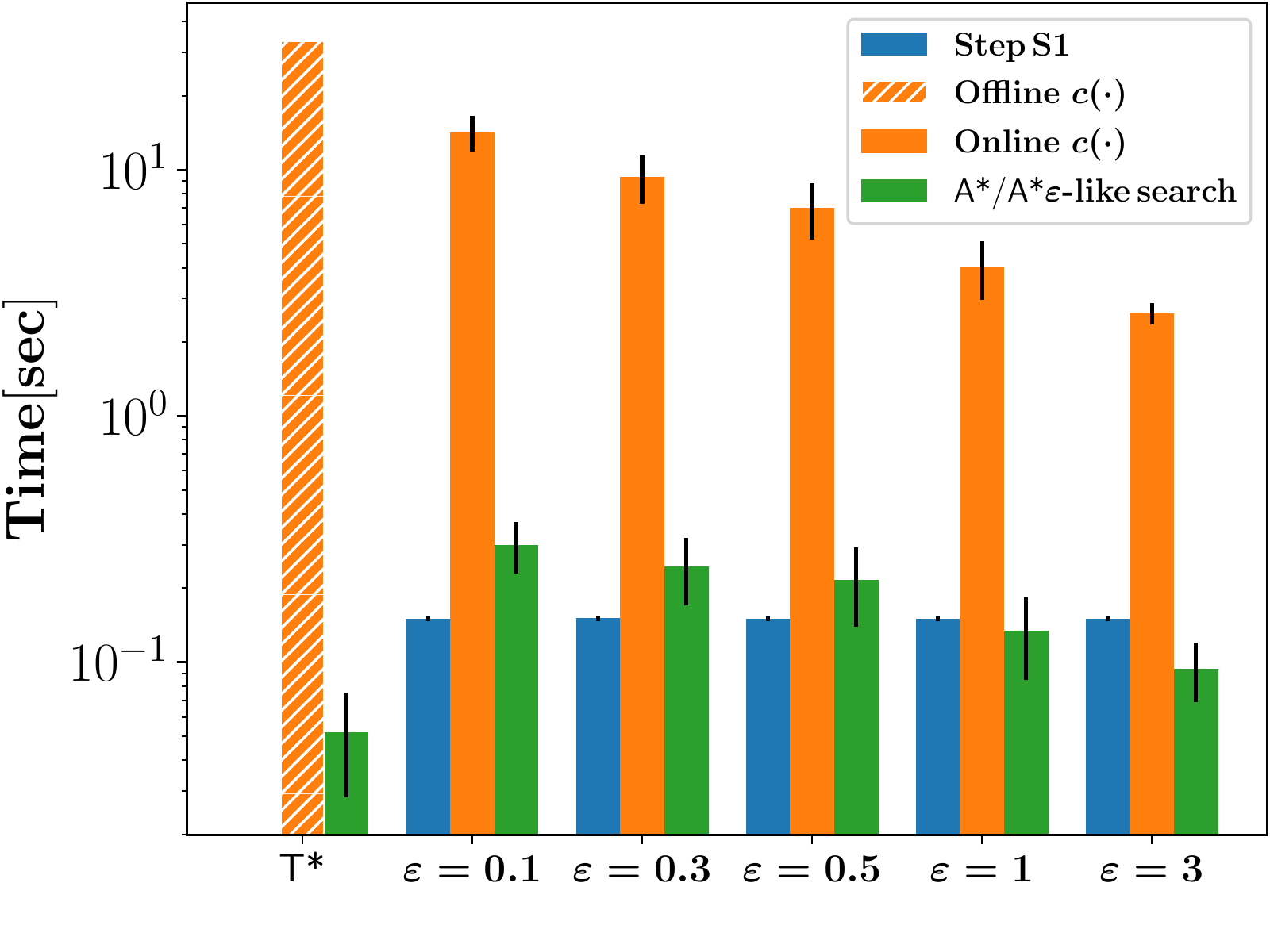}}\,
   \subfloat[\label{fig:no_wind_results-d}]{\includegraphics[height=3.8cm]{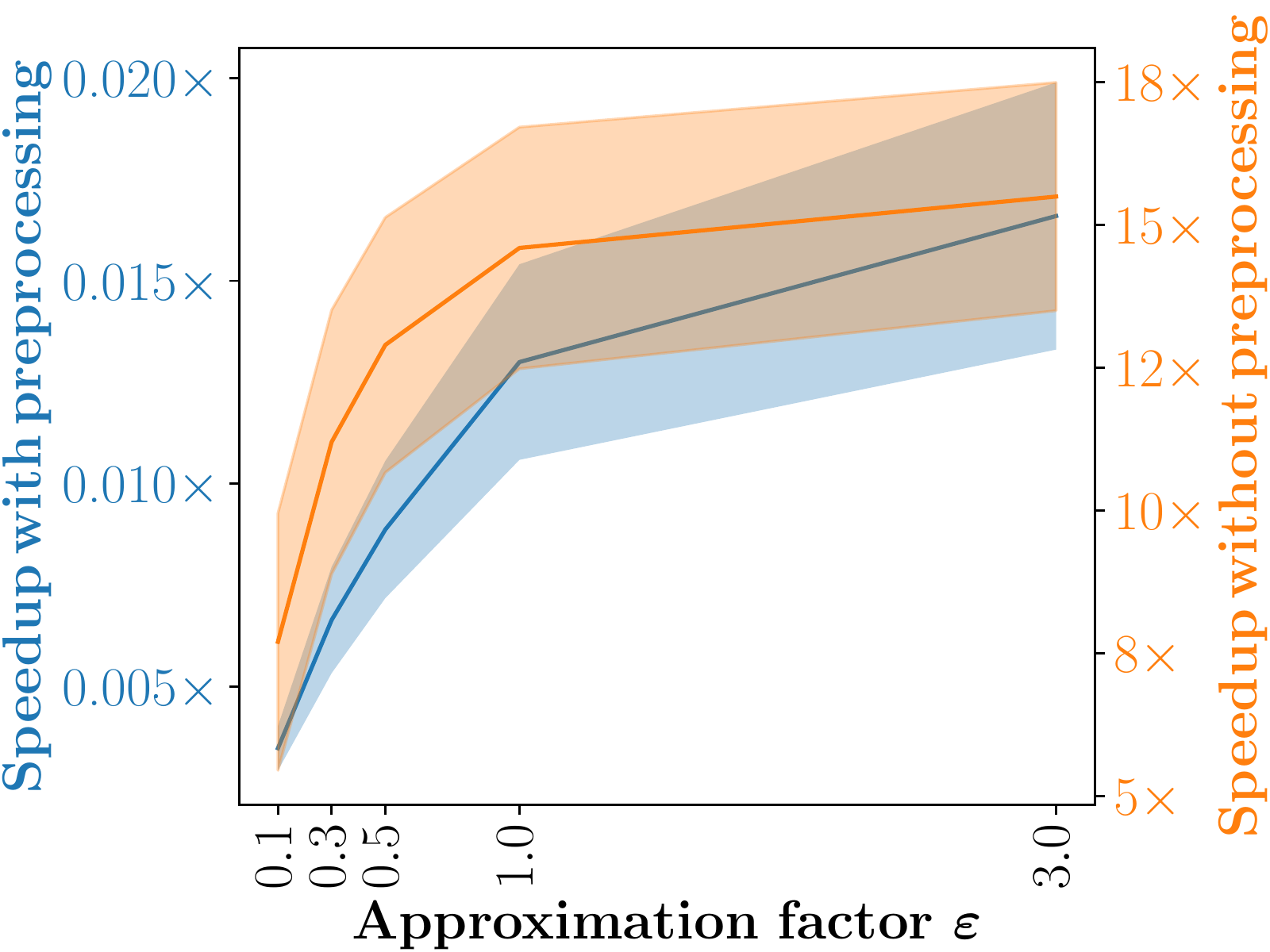}}
   \caption{
   Comparison of \tstar and \tstareps in static environmental conditions over~$100$ randomly chosen~start and~goal configurations.
      \protect\subref{fig:no_wind_results-a} 
      Solution found by~\tstar and by~\tstareps with~$\varepsilon = 3$ for one representative experiment. 
      The beginning of the solution found by both algorithms is identical, while the end differs and is highlighted on the right-hand side.
      %
      \protect\subref{fig:no_wind_results-b} Average solution cost of~\tstareps compared to~\tstar as a function of~\eps.
      \protect\subref{fig:no_wind_results-c} Breakdown of the running time of \tstar and \tstareps for different values of~\eps.
      \protect\subref{fig:no_wind_results-d} Speedup in running time of~\tstareps over~\tstar obtained with and without pre-computing all time-optimal transitions.
    }
   \label{fig:no_wind_results}
   \vspace{-3mm}
\end{figure*}

We compare (Fig.~\ref{fig:lowerBounds}) the three lower bounds w.r.t. the number of time-optimal transitions computed by~\tstareps and its running time.
As expected, the Euclidean distance is the least-informative lower bound, thus resulting in the maximal number of time-optimal transitions computed for all values of~$\varepsilon$. Moreover, even though it is efficient-to-compute, it does a poor job in guiding the search, and thus the runtime is high.
Computing Dubins path while accounting for obstacles is more informative but also more expensive-to-compute. It does result in the smallest number of calls to~$c(\cdot)$ but has longer running times when compared to using Dubins paths that do not account for obstacles.
The latter also has a comparable number of calls to~$c(\cdot)$.
Thus, we use  Dubins paths without obstacles as the lower bound in the rest of the evaluation.


\vspace{-1mm}

\subsection{Motion Planning in Static Environmental Conditions}\label{subsec:eval-static}


{We compare \tstareps with \tstar in static environmental conditions, i.e., where the system dynamics are invariant to the robot's placement and orientation in the scenario. 
%
}
As we can see in Fig.~\ref{fig:no_wind_results-c}, computing time-optimal transitions dominate the running time of all algorithms.
{
If the system dynamics are known in advance, these can be pre-computed in an offline step by \tstar, and it outperforms \tstareps for all $\varepsilon$ by a factor of between~$60\times$ and $290\times$.
However, if we account for this preprocessing time, \tstareps dramatically reduce the run time by a factor ranging between $8\times$ and $15\times$.
It is worth noting that in either case,~\tstareps finds a  solution with an average cost that is much lower than the guaranteed suboptimality bound.}

\subsection{Motion Planning in Dynamic Environmental Conditions}\label{subsec:eval-dynamic}

\begin{figure*}[h]\centering
  \subfloat[\label{fig:wind_results_large_benchmark-a}]{\includegraphics[height=4.2cm]{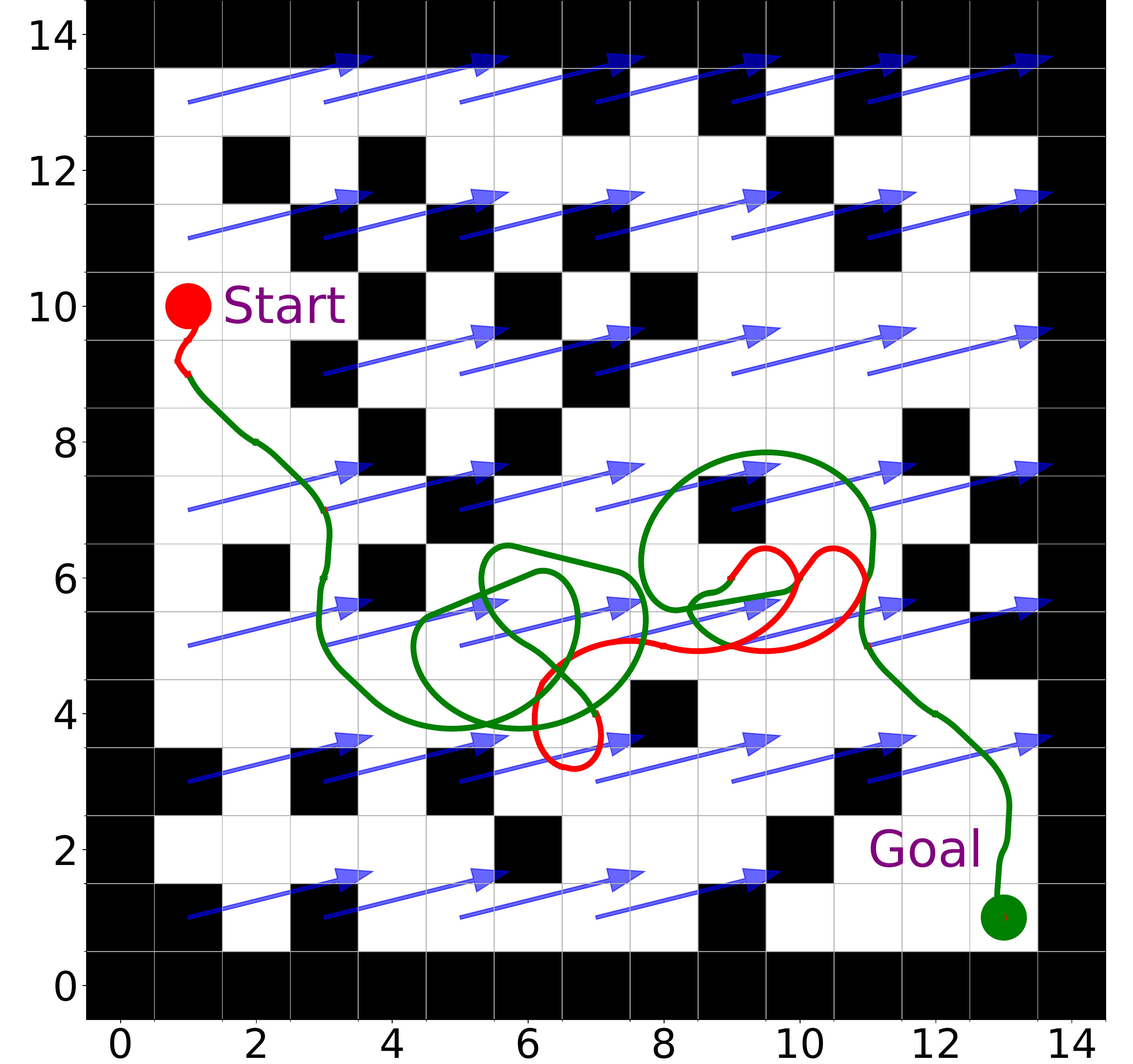}}
  \subfloat[\label{fig:wind_results_large_benchmark-b}]{\includegraphics[height=4.2cm]{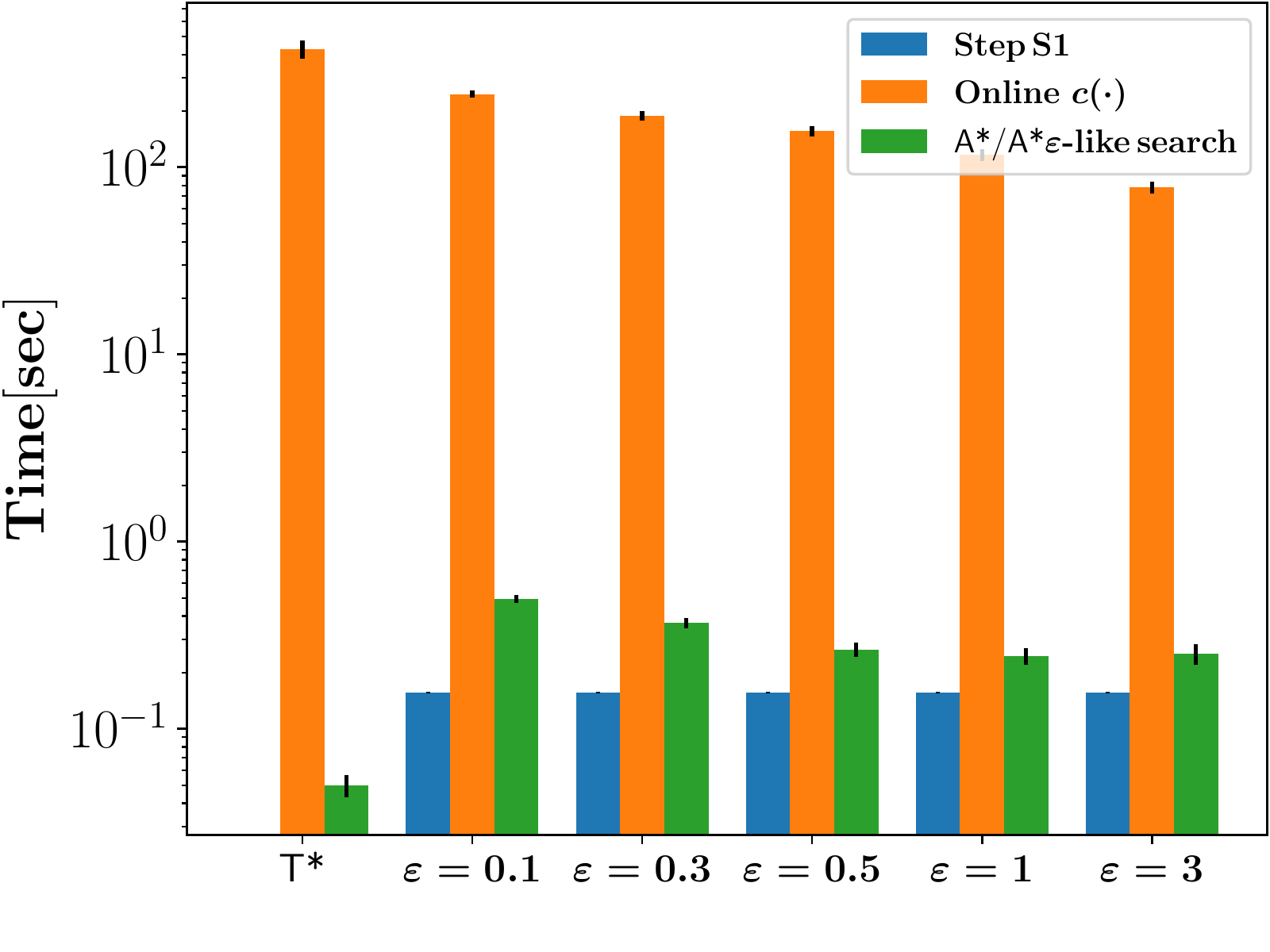}}
  \subfloat[\label{fig:wind_results_large_benchmark-c}]{\includegraphics[height=4.2cm]{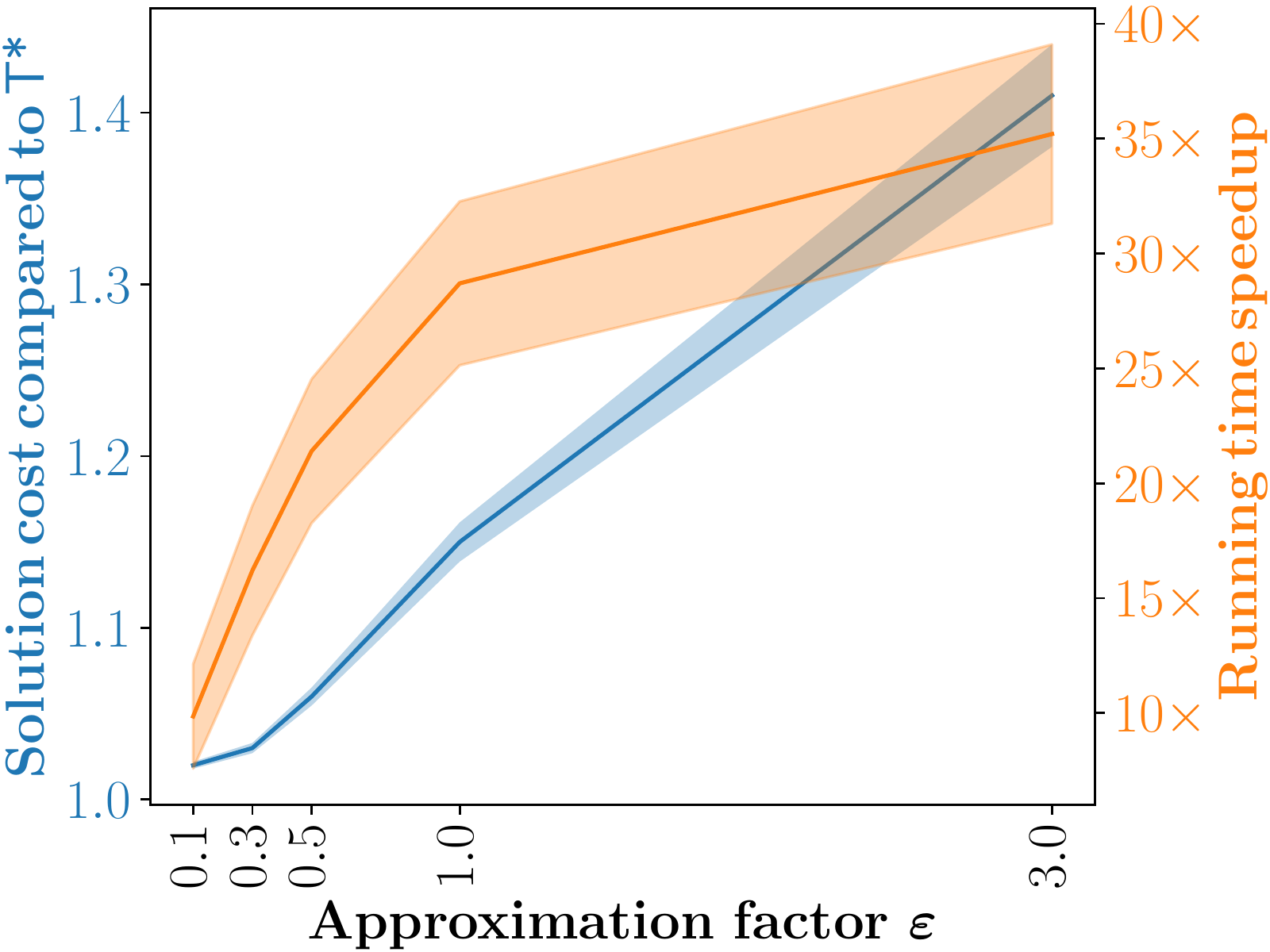}}
  \caption{{
      Comparing~\tstar and~\tstareps in dynamic environmental conditions where the magnitude and direction of wind currents (blue arrows) are given at the query time.
      \protect\subref{fig:wind_results_large_benchmark-a}
      {Representative solution obtained on one randomly-generated map with randomly-generated start and goal configurations, wind conditions and minimum velocity~$v_{\min}$.}
      \protect\subref{fig:wind_results_large_benchmark-b} Running time breakdown of \tstar and \tstareps for the selected values of~\eps.
      \protect\subref{fig:wind_results_large_benchmark-c} 
      Average solution cost compared to the cost of the optimal solution and speedup in running time as a function of~\eps.
      }
  }
  \label{fig:wind_results_large_benchmark}
  \vspace{-3mm}
\end{figure*}
In the previous section, we considered static environmental conditions and showed that using an approximation factor allows reducing \tstareps's planning times dramatically.
However, if the system's dynamics are known in advance, \tstar may perform all its computationally-expensive operations in a preprocessing stage.
In contrast, this is not the case when the system's dynamics are not known in advance.
It can be due to changes in the maximal speed that the system can take (due to, e.g., payload changes of the system or regulatory changes happening just before a query is received) or due to dynamic environmental conditions such as wind currents whose magnitude and direction is known only when a query is received.
We use the latter case to demonstrate the efficacy of~$\tstareps$. In particular, we assume that the magnitude of wind currents is uniform across a given scenario.

It is worth noting that this dynamic setting, analyzed by Techy and Woolsey~\cite{minimum_time_const_wind}, can be seen as a generalization of the setting described in Sec.~\ref{sec:problem} and by Mittal et al.~\cite{Dubins_with_currents}. 
To account for this more-involved dynamic setting, we first note that one cannot use rotation and symmetry between transitions to reduce the total amount of unique transitions. 
Thus, in this setting, the total number of time-optimal transitions is~512.
Moreover, this requires an adaptation of how time-optimal transitions are computed (see the Appendix for a description of the new model). This adaptation, based on the code for computing time-optimal transitions~\cite{Wolek}, is publicly available\footnote{\url{https://github.com/CRL-Technion/Variable-speed-Dubins-with-wind}}.
The second notable change is that computing lower bounds (namely, when calling~$\hat{c}$, requires some care, and the modifications are detailed in the Appendix.

{For dynamic environmental conditions, we used the same randomly generated scenarios as described earlier.
For each such scenario, we generated $10$ instances with varying wind and minimum velocity values. In particular, we sampled wind conditions from $\|\vec{w}\|
\in[0.14,0.41]$ and minimum velocity values $v_{\min}\in [0.4,0.9]$.
In total, we ran $1000$ experiments and the averaged results appear in ~Fig.~\ref{fig:wind_results_large_benchmark}.}
The most computationally demanding component in both~\tstareps and~\tstar is still computing the time-optimal transitions, which requires two orders of magnitude more time than any other algorithmic component. 
However, we observe that the approximation factor allows for a dramatic reduction of the {running time with  little compromise on the quality of the solution.
For example, as it can be seen in Fig~\ref{fig:wind_results_large_benchmark-c}, using~$\varepsilon = 1$, \tstareps guarantees a solution whose cost is at most twice~$c(\gamma^*)$ the cost of the optimal solution, but returns a path whose average cost is no more than~$1.15 \cdot c(\gamma^*)$.
It is done while obtaining a speedup in run time by an average factor of~$28\times$.
In general, as depicted in Fig.~\ref{fig:wind_results_large_benchmark-c},
increasing $\varepsilon$ results in a dramatic running-time improvement at the cost of slightly lower solution quality.}

\subsection{Discussion---Alternative Approaches} \label{subsec:additional_approaches}

Wilson \emph{et al.}~\cite{GMDM} reduced computational effort by allowing the system to follow Dubins paths but using several radii, and evaluated this approach when using three different radii.
This heuristic approach reduces runtime, produces near-optimal solutions but with no guarantees regarding their quality.
From our comparison of~\tstareps and Wilson's model in static environments, we found that running times are faster by several orders of magnitude, but path costs are larger by an average of \SI{18}{\percent}.
However, when evaluating their approach in windy conditions, 
the speedup in running time
is comparable to \tstareps (as local transitions cannot be computed analytically and a computationally expensive root-finding problem needs to be solved~\cite{minimum_time_const_wind}), and the quality of paths remain higher (again, with no formal guarantees).

An alternative approach is to compare our work with the approach by Mittal \emph{et al.}~\cite{Dubins_with_currents} that specifically accounted for efficient computation of paths in dynamic environments.
However, the main objective of their approach is online motion-planning under ocean currents, and no importance was given to the quality of paths.
When comparing \tstareps with their approach on representative environments, the path quality obtained by \tstareps was higher (lower execution times) by a factor of more than $2.7\times$.


\section{Conclusions and Future Work}
In this work, we present~\tstareps, a novel algorithmic framework for efficiently finding bounded-suboptimal solutions for time-optimal motion-planning.
This is done by minimizing the number of computations of time-optimal transitions used by the system.
When one cannot pre-compute the time-optimal transitions in advance, this reduces the planning times by orders of magnitude compared to the state-of-the-art.

Future work includes adapting our work for efficient re-planing (e.g., when the wind changes while the system is in motion or dramatic payload changes during trajectory execution changing the system's constraint).
Here, we plan to adapt LPA*~\cite{KLF04} to account for minimizing the number of transitions used.

Another natural extension is to account for settings when the approximation factor~\eps cannot be determined a priori by the user.
Here, it is natural to use an \emph{anytime}~\cite{any_time_astar} approach.
Instead of providing a fixed~\eps, the user would provide an upper bound for the algorithm's running time.
We suggest starting with an initially high value of the approximation factor~\eps to compute an initial solution quickly and then progressively decreases \eps as the time permits while reusing transitions computed in earlier search episodes.


\vspace{-1mm}

\section*{Appendix}\label{sec:appendix}

\begin{figure}[!b]\centering
  \includegraphics[width=0.65\linewidth]{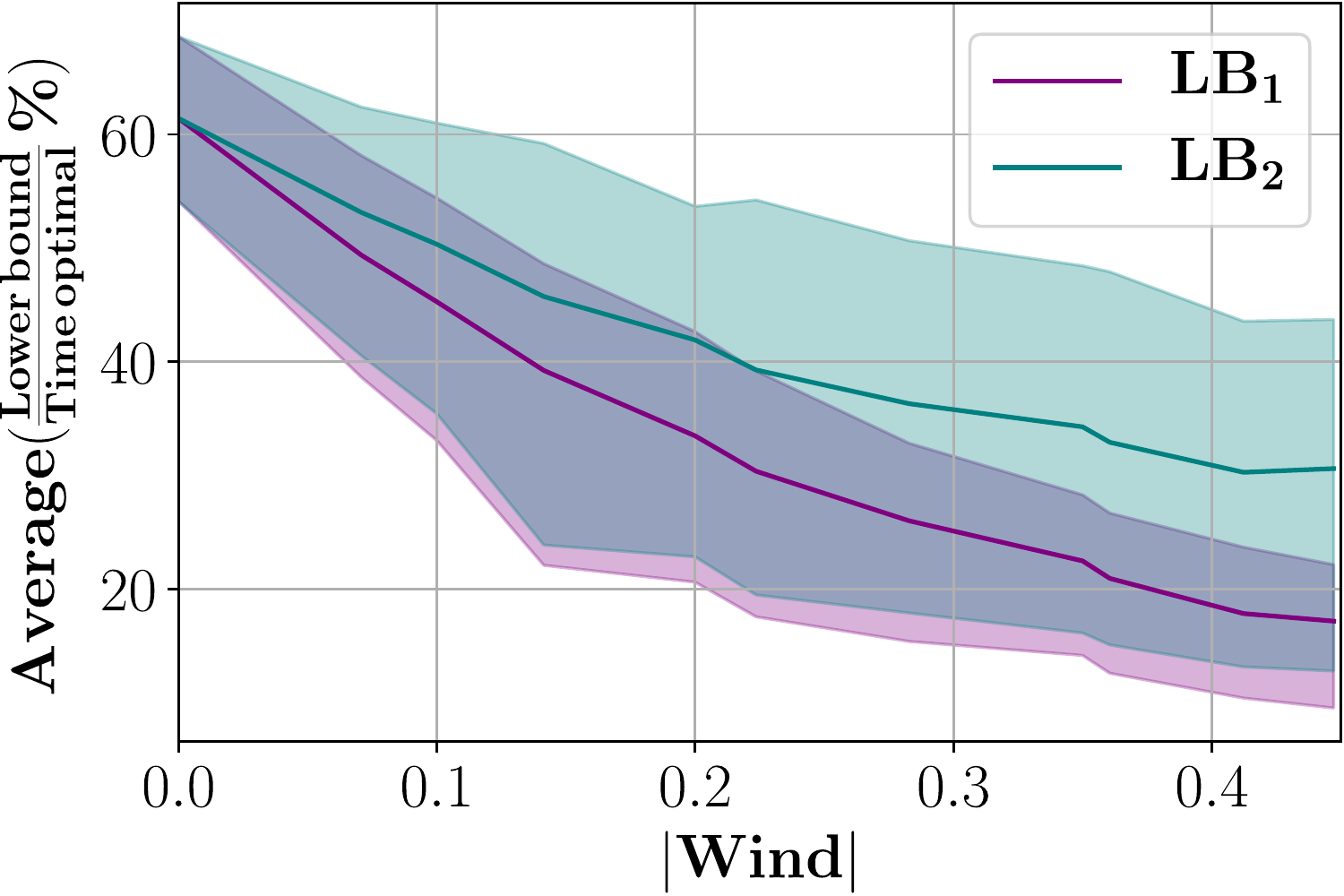}
   \caption{Average ratio between each lower bound and the time-optimal cost as a function of the wind's modulus~(solid lines). 
   Colored regions represent a~$60\%$ non-parametric confidence interval of the true value.
   \label{fig:lower_bound_wind}
   }
   \vspace{-2mm}
\end{figure}
To consider the setting where we need to account for wind currents, we detail the updated dynamics model and present two approaches to compute lower bounds on transitions cost-efficiently.

\subsection{Updated Model}
We assume that there is a constant wind~$\vec{w} = (w_x, w_y)$ that changes the original system dynamics (Eq.~\ref{eq:dynamics}) to
\begin{equation}
\left(\begin{matrix}
    \dot{x}(t)\\
    \dot{y}(t)\\
    \dot{\theta}(t)
\end{matrix} \right)
=
\left(\begin{matrix}
    v(t) \cos\theta(t) + w_x\\
    v(t) \sin\theta(t) + w_y\\
    u(t)
\end{matrix} \right).
\end{equation}

\noindent
Furthermore, we assume that regardless of the magnitude of the wind, the vehicle will move forward. 
It is ensured by the additional assumption that the wind magnitude is smaller than the minimum speed of the system~$v_{\textup{min}}$, i.e.,
\begin{equation}
  \vert \vec{w} \vert < v_{\textup{min}}.
  \label{eq:windlimit}
\end{equation}

\subsection{Updated Lower Bounds}
The wind component may significantly influence the turning capabilities (in the reference frame of the ground). 
The smallest turning radius is obtained when flying directly against the wind.
We denote it by~$\rho_\text{LB}$, the lower bound radius, and have that
\begin{equation}
  \rho_\text{LB} = \left(1-\frac{\vert \vec{w} \vert}{v_{\textup{min}}}\right) \, \rho _{\min}.
  \label{r_min_wind}
\end{equation}
To this end, a lower bound on the true length of the path between two configurations~$s_1$ and~$s_2$ can be determined based on the length~$\mathcal{L}(s_1, s_2, \rho_\text{LB})$ of Dubins path with the lower bound radius~$\rho_\text{LB}$.
Now, if this value is divided by an upper bound on the effective maximal speed~$v_{\max}^{\text{effective}}$ in the reference frame of the ground, it provides a lower bound on the true cost (i.e., the travel time).

We suggest two lower bounds based on how~$v_{\max}^{\text{effective}}$ is determined.
The first approach assumes that~$v_{\max}^{\text{effective}} = v _{\max} + \vert \vec{w} \vert$.
Namely, the effective maximal speed is the sum of the system's maximum speed and the wind magnitude.
Thus, the first lower bound~$\text{LB}_1$ is defined as 
\begin{equation}
   \text{LB}_1(s_1, s_2) = \frac{\mathcal{L}(s_1, s_2, \rho_\text{LB})}{v _{\max} + \vert \vec{w} \vert}.
   \label{LB1}
\end{equation}

The second lower bound~$\text{LB}_2$ considers the direction between the  start and end configurations ($s_1$ and~$s_2$, respectively) by computing the ground speed~$\vec{v_{\textup{G}}}$ in the given direction~$\vec{d}$.
The direction is determined based on the configurations~$s_1$ and~$s_2$ as~$\vec{d} = \left( {s_{2}^\text{xy} - s_{1}^\text{xy}} \right) / \left({\vert s_{2}^\text{xy} - s_{1}^\text{xy} \vert} \right)$,
where~$s_{i}^\text{xy}$ stands for the~$x,y$ coordinates of the configuration~$s_i$.
The ground speed~$\vec{v_{\textup{G}}}$ vector can be expressed by the system's aerial speed~$\vec{v}$ and wind~$\vec{w}$ as $\vec{v_{\textup{G}}} = \vec{v} + \vec{w}$.

Following Eq.~\ref{eq:windlimit}, the maximum ground speed occurs at the system's maximum speed, i.e.,~$\vert \vec{v} \vert = v_\text{max}$.
The angle between~$\vec{v_{\textup{G}}}$ and~$\vec{w}$ can be written as{~$\cos \angle (\vec{v_{\textup{G}}}, \vec{w}) = \frac{\vec{d} \cdot \vec{w}}{\vert \vec{w} \vert}.$}
Subsequently, the ground speed can be expressed using the cosine rule as~$\vert \vec{v_{\textup{G}}} \vert = \vec{d} \cdot \vec{w} + \sqrt{( \vec{d} \cdot \vec{w} )^2 + v _{\max}^2 -\vert \vec{w} \vert ^2 }.$
Finally, the second lower bound~$\text{LB}_2$ is defined as
\begin{equation}
  \text{LB}_2(s_1, s_2) = \frac{\mathcal{L}(s_1, s_2, \rho_\text{LB})}{\vert \vec{v_{\textup{G}}} \vert }.
  \label{LB2}
\end{equation}
     
Fig.~\ref{fig:lower_bound_wind} shows the average ratio between each lower bounds and the true cost as a function of the wind force. 
For both lower bounds, their estimation of the true cost becomes looser as the wind force increases.
Notice that~$\text{LB}_2$ is more informative than~$\text{LB}_1$ because accounting for the direction enables reducing the upper bound on the ground speed. 
In addition, notice that both lower bounds reduce to the system's minimum-speed Dubins path divided by the maximum speed in static environmental conditions.
Thus, we used~$\text{LB}_2$ as lower bound for dynamic environmental conditions.

\bibliographystyle{IEEEtran}
\bibliography{main}

\end{document}